\documentclass[doublecolumn,conference,letterpaper]{IEEEtran}
\usepackage[utf8]{inputenc} 
\usepackage[T1]{fontenc}
\usepackage{graphicx}
\usepackage{caption}
\usepackage{subcaption}  
\usepackage{url}              
\usepackage{ifthen}          
\usepackage{cite}             
\usepackage[cmex10]{amsmath}  
\usepackage{amssymb}
\usepackage{amsthm}
\usepackage{mleftright}       
\mleftright                   
\usepackage{booktabs}         
\usepackage{xcolor}
\usepackage{amsmath,amsfonts}
\usepackage{dsfont}
\usepackage{graphicx}
\newtheorem{proposition}{Proposition} 
\newtheorem{assumption}{Assumption}  
\newtheorem{corollary}{Corollary}  
\newtheorem{lemma}{Lemma}
\newtheorem{theorem}{Theorem}

\newcommand{\ntr}{n_{ \text{tr} }}

\begin{document}
	
	\title{Generalization and Informativeness \\of Conformal Prediction} 
	
	\author{%
  \IEEEauthorblockN{Matteo Zecchin, Sangwoo Park, Osvaldo Simeone}
  \IEEEauthorblockA{Centre for Intelligent Information Processing Systems\\Department of Engineering\\
                    King’s College London\\
                     London, United Kingdom\\
                    Email: \{matteo.1.zecchin,sangwoo.park,osvaldo.simeone\}@kcl.ac.uk}
  \and
  \IEEEauthorblockN{Fredrik Hellström}
  \IEEEauthorblockA{
                    Centre for Artificial Intelligence\\
                    Department of Computer Science\\
                    University College London\\ 
                    London, United Kingdom\\
                    Email: f.hellstrom@ucl.ac.uk}
}

	\maketitle
	
	\begin{abstract}
	The safe integration of machine learning modules in decision-making processes hinges on their ability to quantify uncertainty. A popular technique to achieve this goal is conformal prediction (CP), which transforms an arbitrary base predictor into a set predictor with coverage guarantees. While CP certifies the predicted set to contain the target quantity with a user-defined tolerance, it does not provide control over the average size of the predicted sets, i.e., over the informativeness of the prediction. In this work, a theoretical connection is established between the generalization properties of the base predictor and the informativeness of the resulting CP prediction sets. To this end, an upper bound is derived on the expected size of the CP set predictor that builds on generalization error bounds for the base predictor. The derived upper bound provides insights into the dependence of the average size of the CP set predictor on the amount of calibration data, the target reliability, and the generalization performance of the base predictor. The theoretical insights are validated using  simple numerical regression and classification tasks.
	\end{abstract}
	\section{Introduction}
 
    \emph{Context and motivation}: In safety-critical domains such as health, finance, and engineering  \cite{beam2018big,goodell2021artificial,hewing2020learning}, machine learning models are typically required to provide accurate estimates of the uncertainty associated with their outputs \cite{lu2022fair,wisniewski2020application,zecchin2023forking}.  \emph{Conformal prediction} (CP) offers a practical solution to produce certified ``error bars'' in the form of set predictions obtained by post-processing the outputs of a fixed, pre-trained, base predictor \cite{vovk2005algorithmic}. With CP, uncertainty is captured by the size of the predicted sets, with smaller sets providing more informative predictions.  While CP is guaranteed to meet a user-defined coverage level, the informativeness of the predicted set can usually only be assessed at test time. 
\begin{figure}
    \centering
    \begin{subfigure}{.24\textwidth}
      \centering
      \includegraphics[width=\linewidth]{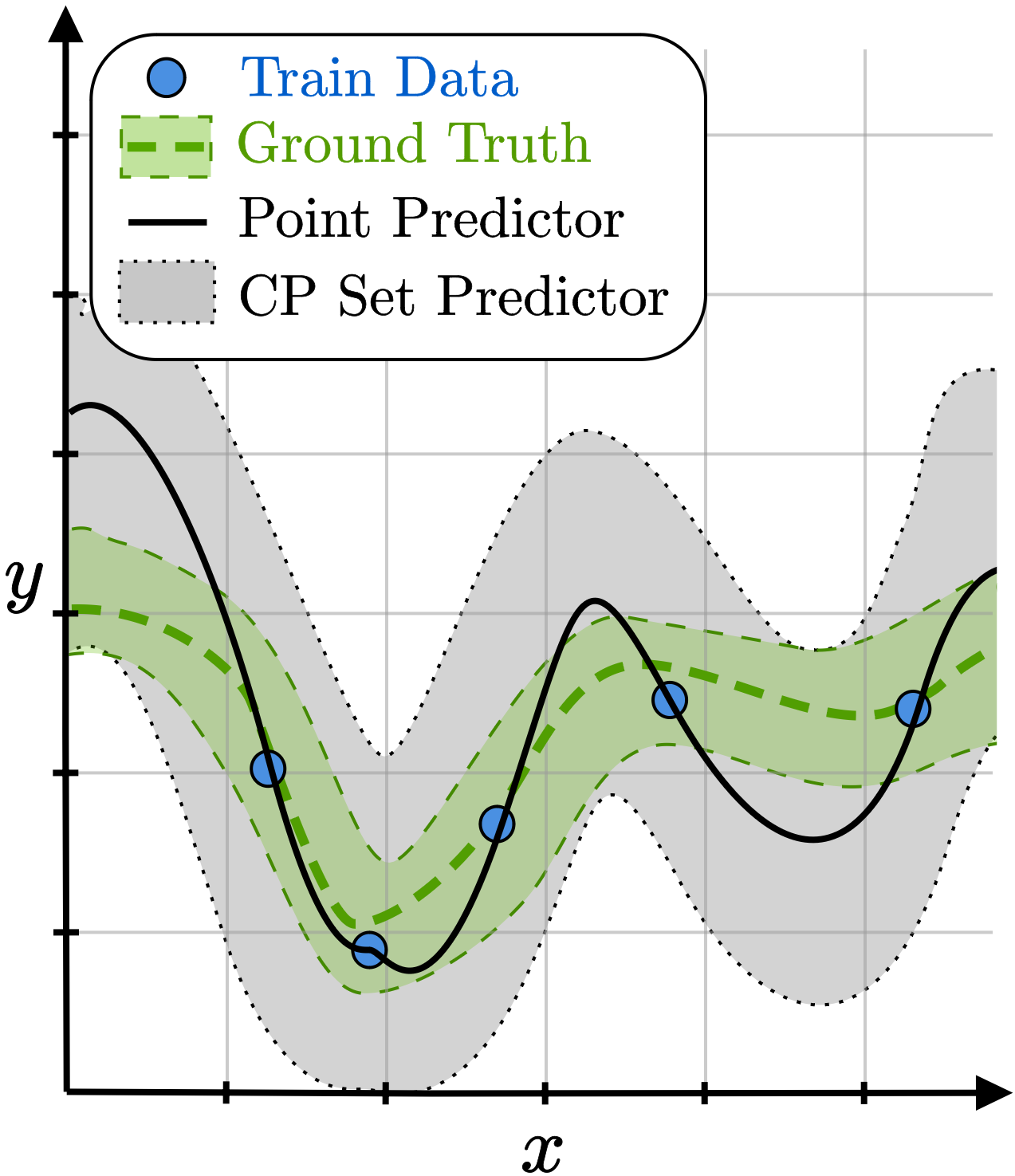}
      \label{fig:sub1}
    \end{subfigure}%
    \hfill
    \begin{subfigure}{.24\textwidth}
      \centering
      \includegraphics[width=\linewidth]{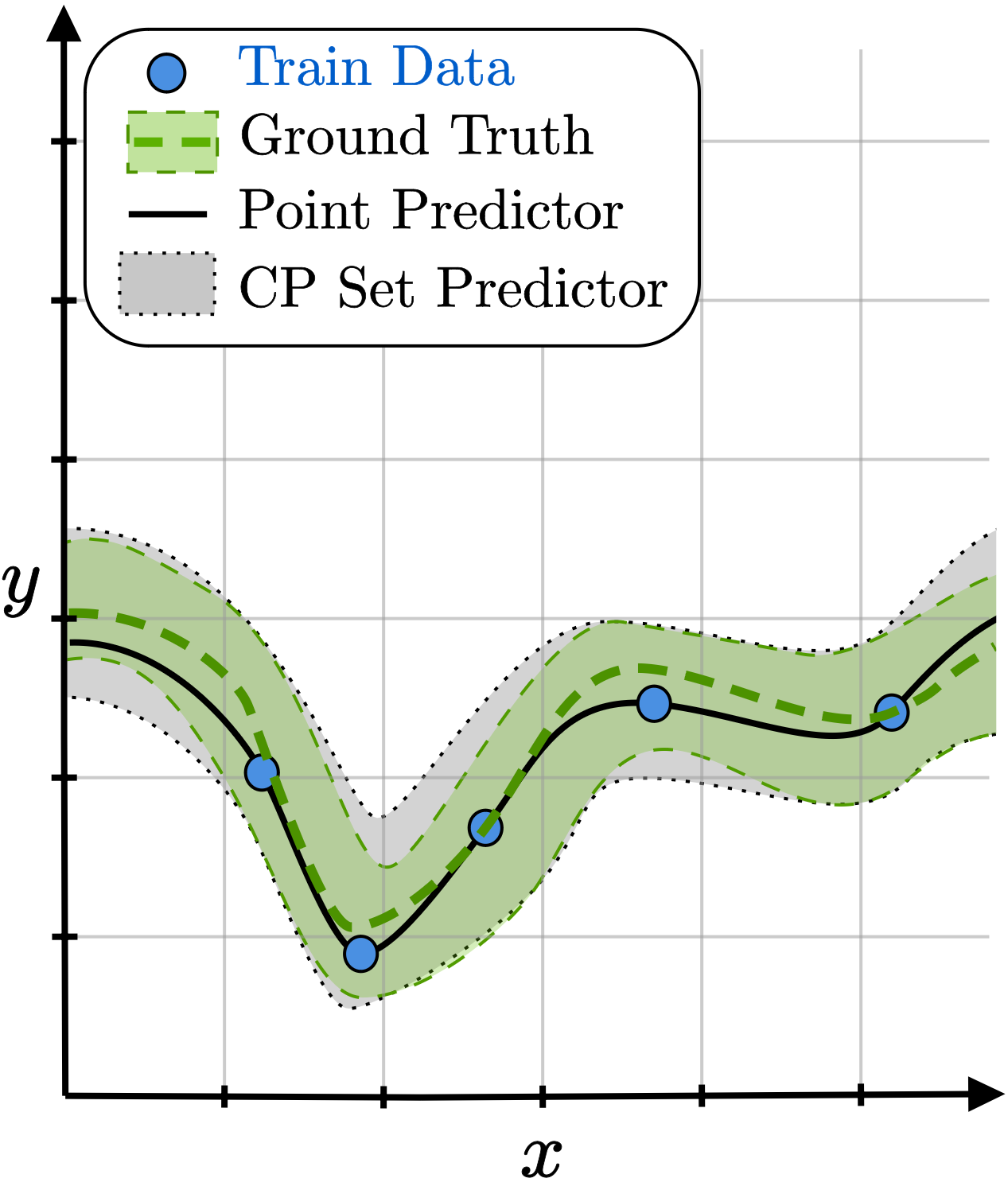}
      \label{fig:sub2}
    \end{subfigure}
    \vspace{-1em}
    \caption{Conformal prediction (CP) set predictors (gray areas) obtained by calibrating a base predictor with a higher generalization error on the left and a lower generalization error on the right. Thanks to CP, both set predictors satisfy a user-defined coverage guarantee, but the inefficiency, i.e., the average prediction set size, is larger when the generalization error of the base predictor is larger. }
     \vspace{-0.7em}
    \label{fig:illustrative}
    \end{figure}

     The most practical version of CP, known as \emph{inductive CP}, splits the available data into a training set and a calibration set. Training data are used to optimize an arbitrary base model, while the calibration data are leveraged to determine the ``error bars'' around the decisions produced by the base model. As illustrated in Figure \ref{fig:illustrative}, a more accurate base predictor, providing better generalization outside the training set, tends to yield smaller prediction sets upon the application of CP. The goal of this work is to connect the generalization properties of the base predictor with the informativeness of the set predictions produced by CP.

    \noindent \emph{Related work}: CP offers distribution-free marginal guarantees of coverage at any user-defined level \cite{vovk2005algorithmic,angelopoulos2021gentle}. The informativeness of the set predictors produced by CP depends on the quality of the base predictor and on the choice of criterion, known as non-conformity score, used to determine the prediction sets \cite{kato2023review}. In the asymptotic regime of a large training and calibration data set, it is known that CP sets produced using nonparametric density estimators are optimal, matching the set produced by an oracle that knows the data distribution \cite{lei2013distribution,lei2014distribution,sadinle2019least}. In the finite-sample regime of interest in this work, a recent paper has provided methods for estimating the expected predicted set size based on additional held-out data \cite{dhillon2023expected}. In contrast, our work derives a bound relating the expected set size to the generalization performance of the base predictor.

    A recent contribution has combined PAC-Bayes generalization theory \cite{alquier2021user,hellstrom2023generalization} and CP to obtain generalization bounds on the coverage and informativeness of the set predictions \cite{sharma2023pacbayes}. Similarly to our work, the authors derive an upper bound on the expected prediction set size. However, their focus is on deriving a design criterion for a modified CP procedure that jointly leverages training and calibration data for enhanced efficiency. Unlike \cite{sharma2023pacbayes}, our results pertain to the conventional inductive CP approach.

  \noindent \emph{Main contributions and organization}: In this paper, we provide an upper bound on the expected size of CP predictions that depends on the generalization error of the base predictor. The derived bound provides insights into the  dependence of the CP-based predicted set size on the dimension  of the training set, on the amount of calibration data, and on the target reliability level.  The theoretical results are validated on simple classification and regression tasks. 

  The rest of the paper is organized as follows. Section II describes the problem, Section III presents the main result, and Section IV describes two numerical examples, with Section V concluding the paper.

	\section{Problem Definition}
	We consider the problem of characterizing the informativeness of  set predictors obtained using the CP framework \cite{vovk2005algorithmic,angelopoulos2021gentle}. To this end, we study the standard supervised learning setting in which each data point $Z=(X,Y)\in \mathcal{X}\times\mathcal{Y}$ consists of an input feature $X\in\mathcal{X}$ and a label  $Y\in\mathcal{Y}$, which are jointly distributed according to an   \emph{unknown} distribution $P_Z=P_{XY}$. Throughout, we use capital letters for random variables and the corresponding lowercase letters for realizations. 
 
 In this context, a \emph{set predictor} $\Gamma$ maps an input feature $x$ into a subset of the label space $\Gamma(x)\subseteq\mathcal{Y}$. For a given reliability level $\alpha$, a set predictor $\Gamma$ is said to be \emph{$\alpha$-reliable} if it includes the true label $Y$ with a probability no smaller than $1-\alpha$, i.e., 
	\begin{align}
		\Pr[Y\in\Gamma(X)]\geq 1-\alpha.
		\label{eq:cov_guar_gen}
	\end{align} Note that the coverage guarantee \eqref{eq:cov_guar_gen} is marginal in the sense that it applies on average with respect to the jointly distributed input variable $X$ and output variable $Y$.
 
	The coverage guarantee \eqref{eq:cov_guar_gen} can always be satisfied by a set predictor that returns the entire label space $\mathcal{Y}$ for any input feature $X$. Therefore, for a target coverage level $\alpha$, a key metric to rank a set predictor $\Gamma$ is its \emph{inefficiency}, which is defined as the average predicted set size
	\begin{align}
		\Lambda(\Gamma)=\mathbb{E}\left[|\Gamma(X)|\right].
		\label{eq:set_size}
	\end{align}
	
	Given access to a set of $n$ i.i.d data samples $\mathcal{D}=\{Z_i\}^n_{i=1}\sim P^{\otimes n}_{Z}$, a common practice to produce a \emph{reliable} and \emph{efficient}, or informative, set predictor consists in training a machine learning model on a subset of the data set $\mathcal{D}_{\text{tr}}$, and then to use the remaining part of the data $\mathcal{D}_{\text{cal}}=\mathcal{D}\setminus \mathcal{D}_{\text{tr}}$ to \emph{calibrate} the resulting model using CP. 
 We denote the sizes of the training data set $\mathcal{D}_{\text{tr}}$ and of the calibration data set $\mathcal{D}_{\text{cal}}$ as $n_{\text{tr}}$ and $n_{\text{cal}}=n-n_{\text{tr}}$, respectively.
	
	\subsection{Model Training}
	
	The first step towards the definition of a reliable set predictor is the training of a base point predictor. To this end, we assume that the learner selects a  \emph{model class}  $\mathcal{F}=\{f_{\theta}: \theta \in \Theta \}$, where the prediction function $f_{\theta}:\mathcal{X}\to\hat{\mathcal{Y}}$, such as a neural network, is parameterized by a vector $\theta\in\Theta$. 
 
 A \emph{learning algorithm} defines a mapping from the training data set  $\mathcal{D}_{\text{tr}}$ to a probability $ Q(\theta|\mathcal{D}_{\text{tr}})$ over the space of model parameters $\Theta$. The distribution $ Q(\theta|\mathcal{D}_{\text{tr}})$ may be implicit, as in the case of randomized algorithms such as stochastic gradient descent. Given a distribution $Q(\theta|\mathcal{D}_{\text{tr}})$, a decision on an input $X$ is made by first drawing a model $\theta\sim Q(\theta|\mathcal{D}_{\text{tr}})$ and then using the prediction $\hat{Y}=f_\theta(X)$ (see, e.g., \cite[Chapter 12] {simeone2022machine}).

	An important metric to gauge the performance of a model class $\mathcal{F}$ and of a training algorithm $Q(\theta|\mathcal{D}_{\text{tr}})$ under the data distribution $P_Z$ is the \emph{generalization error}. The generalization error measures the discrepancy between the training and testing performance of the learning algorithm $Q(\theta|\mathcal{D}_{\text{tr}})$. The generalization error is a function of the training algorithm $Q(\theta|\mathcal{D}_{\text{tr}})$, viewed as a function of the training data set $\mathcal{D}_{\text{tr}}$, as well as of the  training data set $\mathcal{D}_{\text{tr}}$  \cite{alquier2021user,hellstrom2023generalization}.
	
	\subsection{Reliable Set Predictors via Conformal Prediction}

	CP is a post-hoc calibration technique that converts a point predictor $f_\theta:\mathcal{X}\to\hat{\mathcal{Y}}$ into an $\alpha$-reliable set predictor $\Gamma_\theta$. Fix a bounded function $R:\hat{\mathcal{Y}}\times \mathcal{Y}\to [0,R_{\text{max}}]$ to measure the loss of the predictor, e.g., the  0-1 loss or the squared loss. This is known as the \emph{non-conformity (NC) scoring function}. 
 For each $i$-th sample  in the calibration data set $\mathcal{D}_{\text{cal}}$, we define the $i$-th \emph{calibration NC score}, for $i=1,\dots,n_{\text{cal}}$, as 
	\begin{align}
		R_i=R(f_{\theta_i}(X_i),Y_i), \quad \text{where } \theta_i\sim Q(\theta|\mathcal{D}_{\text{tr}}).
		\label{eq:cal_NC_score}
	\end{align} 
	By \eqref{eq:cal_NC_score}, the calibration score is obtained by first sampling a model parameter  $\theta_i\sim Q(\theta|\mathcal{D}_{\text{tr}})$ to determine a predictor $f_{\theta_i}$, and then evaluating the NC score $R_i$ of the prediction $f_{\theta_i}(X_i)$ for the calibration data point $Z_i=(X_i,Y_i)$.
	
	We denote the collection of the calibration NC scores as the set $\mathcal{R}_{\text{cal}}=\{R_i\}^{n_{\text{cal}}}_{i=1}$, and  the \emph{empirical calibration cumulative distribution function} (c.d.f.) of the calibration NC scores as 
    \begin{align}
        \hat{F}_{\theta}(r|\mathcal{D}_{\text{cal}})=\frac{1}{n_{\text{cal}}}\sum^{n_{\text{cal}}}_{i=1}\mathds{1}\{R(f_{\theta_i}(x_i),y_i)\leq r\}.
        \label{eq:cal_cdf}
    \end{align} We also let $\mathcal{Q}_{1-\alpha}(\mathcal{R}_{\text{cal}})=\hat{F}_{\theta}^{-1}((n_\alpha+1)/n_{\text{cal}}|\mathcal{D}_{\text{cal}})$ denote the $(n_\alpha+1)$-th largest element of the calibration NC scores, where we define 
 \begin{align}
     n_\alpha=\lceil(n_{\text{cal}}+1)(1-\alpha)\rceil-1,
 \end{align} and we used  the  notation $\hat{F}_{\theta}^{-1}(p|\mathcal{D}_{\text{cal}})$ to represent the smallest value of $r$ such that the inequality $\hat{F}_{\theta}(r|\mathcal{D}_{\text{cal}})\geq p$ holds.

	For a test input feature $X$, and given the randomly drawn model $\theta\sim Q(\theta|\mathcal{D}_{\text{tr}})$, the CP set predictor is obtained by including all values of $y\in\mathcal{Y}$ with a NC score that is no larger than  $\mathcal{Q}_{1-\alpha}(\mathcal{R}_{\text{cal}})$, i.e., \cite{vovk2005algorithmic}
	\begin{align}
		\Gamma^{\text{CP}}_{\theta}(X|\mathcal{D}_{\text{cal}})=\left\{y\in\mathcal{Y}: R(f_\theta(X),y)\leq \mathcal{Q}_{1-\alpha}(\mathcal{R}_{\text{cal}})\right\}. 
		\label{eq:cp_predictor}
	\end{align} The CP set predictor \eqref{eq:cp_predictor} is known to be $\alpha$-reliable in the sense that the inequality
	\begin{align}
		\Pr[Y\in\Gamma^{\text{CP}}_{\theta}(X|\mathcal{D}_{\text{cal}})|\mathcal{D}_{\text{tr}}]\geq 1-\alpha
		\label{eq:cov_guar}
	\end{align}
    holds, where the probability is taken over the joint distribution of the model $\theta\sim Q(\theta|\mathcal{D}_{\text{tr}})$, the calibration data $\mathcal{D}_{\text{cal}}\sim P_Z^{\otimes n_{\text{cal}}}$, and the test data $Z\sim P_Z$. Note that the probability \eqref{eq:cov_guar} is conditioned on the training data set, and that the random variables $\theta$, $\mathcal{D}_{\text{cal}}$ and $Z$ are conditionally independent given $\mathcal{D}_{\text{tr}}$.
	Using \eqref{eq:set_size}, we define as 
	\begin{align}
		\Lambda^{\text{CP}}(Q|\mathcal{D}_{\text{tr}})=\mathbb{E}\left[|\Gamma^{\text{CP}}_\theta(X|\mathcal{D}_{\text{cal}})|\right]
        \label{eq:set_size_CP}
	\end{align}
    the inefficiency of the CP set predictor, where the average is taken over the same distribution as in \eqref{eq:cov_guar}.

\subsection{Size of an NC Score}
    
	Let us denote as $P_{R|Y=y,\theta}$ the probability density function (p.d.f) of the NC score $R(f_\theta(X),y)$ for a given model $\theta$. Note that the distribution depends solely on the conditional distribution $P_{X|Y=y}=P_{X,Y=y}/P_{Y=y}$ of the data. Therefore, we can interpret $P_{R|Y=y,\theta}$ as the fraction of inputs for which the score assigned by the model $\theta$ to the label $y$ equals $r$. By averaging over the trained model $\theta\sim Q(\theta|\mathcal{D}_{\text{tr}})$ and over a uniformly selected label $Y\sim\mathcal{U}(\mathcal{Y})$, we obtain the \emph{size of an NC score level  $r$} as \cite{dhillon2023expected}
    \begin{align}
		\gamma(r|Q,\mathcal{D}_{\text{tr}})=\frac{1}{|\mathcal{Y}|}\int_{\mathcal{Y}}  \mathbb{E}_{ Q(\theta|\mathcal{D}_{\text{tr}})}\left[P_{R|Y=y,\theta}(r)\right]\mathrm{d}y.
		\label{eq:exp_mf}
	\end{align} By definition \eqref{eq:exp_mf}, the size of the score $r$ represents the fraction of data points, drawn as explained, that are associated with an NC score level $r$. We make the following assumption. \begin{assumption}
		\label{ass:nc}
		The factor $\gamma(r|Q,\mathcal{D}_{\text{tr}})$ is non-decreasing in $r$ for any data set $\mathcal{D}_{\text{tr}}$.
	\end{assumption}
	
  This assumption holds for common NC scores such as the $\ell_p$ error for regression with a bounded target domain $\mathcal{Y}=[B_l,B_u]$, for which we have  \cite[Table~1]{dhillon2023expected}
	\begin{align}\label{eq:scoreregr}
	    \gamma(r|Q,\mathcal{D}_{\text{tr}})=\frac{2r^{\frac{1}{p}-1}}{p(B_u-B_l)} ,
	\end{align} 
    as well as for classification with the 0-1 loss as
    \begin{align}\label{eq:scoreclass}
	    \gamma(r|Q,\mathcal{D}_{\text{tr}})=\begin{cases}
	        \frac{1}{|\mathcal{Y}|}, &\text{ if } r=0,\\
            1-\frac{1}{|\mathcal{Y}|}, &\text{ if } r=1.
	    \end{cases}
	\end{align}

	\section{Inefficiency and Generalization}
	In this section, we study the interplay between the efficiency of the CP set predictor, as measured by the average set size $\Lambda^{\text{CP}}(Q|\mathcal{D}_{\text{tr}})$ in \eqref{eq:set_size_CP}, and the generalization performance of the training algorithm $Q(\theta|\mathcal{D}_{\text{tr}})$.
    \subsection{Main Result}
    \label{sec:main_res}
	To start, we define a counterpart of the c.d.f. (\ref{eq:cal_cdf}) evaluated using the training data set. As we will see, this will allow us to relate the generalization performance of the training algorithm with the size of the predicted set. Accordingly, the \emph{empirical training c.d.f.} of the NC scores at level $r$ is defined as 
	\begin{align}
		\hat{F}(r|Q,\mathcal{D}_{\text{tr}})=\frac{1}{n_{\text{tr}}}\sum_{i=1}^{n_{\text{tr}}}\Pr\left[R(f_{\theta}(x_i),y_i)< r|\mathcal{D}_{\text{tr}}\right],
		\label{eq:train_NCscores}
	\end{align}
	where the probability is evaluated with respect to the random model $\theta\sim Q(\theta|\mathcal{D}_{\text{tr}})$ for a fixed training data set $\mathcal{D}_{\text{tr}}$. We can view the training c.d.f.  (\ref{eq:train_NCscores}) as an approximation of the population c.d.f. 
	\begin{align}
		F(r|Q,\mathcal{D}_{\text{tr}})=\Pr\left[R(f_{\theta}(X),Y)< r\right],
		\label{eq:true_pop_cdf}
	\end{align} 
	where the probability is computed over the model  $\theta\sim Q(\theta|\mathcal{D}_{\text{tr}})$ and the sample $(X,Y)\sim P_Z$.
	We measure the generalization properties of the training-based estimate \eqref{eq:train_NCscores} via the metric  
		\begin{align}
	\Delta(Q|\mathcal{D}_{\text{tr}})=\sup_{r\in[0,R_{\text{max}}]} \left|F(r|Q,\mathcal{D}_{\text{tr}})\hspace{-0.2em}-\hspace{-0.2em}\hat{F}(r|Q,\mathcal{D}_{\text{tr}})\right|.
	\label{eq:cdf_errror}
	\end{align}

	\begin{assumption}
	\label{ass:gen_bound}
	For any $\delta\!\in\!(0,1)$, there exists a function $\beta (\delta,n_{\text{tr}})$, which satisfies $\beta (\delta,n_{\text{tr}})= O(\log( n_{\text{tr}}))$ for any fixed $\delta$, such that the inequality 
	\begin{align}\label{eq:assum_gen_bound}
		\Delta(Q|\mathcal{D}_{\text{tr}})\leq \frac{\beta(\delta,n_{\text{tr}}) }{\sqrt{n_{\text{tr}}}}
	\end{align}
	holds with probability no smaller than $1-\delta$ with respect to the distribution of the training data set $\mathcal{D}_{\text{tr}}$.
\end{assumption}
Note that, due to the supremum in \eqref{eq:cdf_errror}, the inequality in \eqref{eq:assum_gen_bound} differs from standard generalization bounds. However, as shown formally in Appendix A, Assumption \ref{ass:gen_bound} holds for many practical learning algorithms, such as Gibbs posteriors, stochastic gradient Langevin dynamics (SGLD), and differentially private algorithms. Appendix A also reports an explicit expression for the function $\beta(\delta,n_{\text{tr}})$.

Assumption \ref{ass:gen_bound} stipulates that the approximation error (\ref{eq:cdf_errror}) of the model $Q(\theta|\mathcal{D}_{\text{tr}})$ meets the standard statistical error scaling of $O(1/\sqrt{n_{\mathrm{tr}}})$, neglecting logarithmic terms. As reflected by the main result of this work, summarized in the following theorem, this generalization error can be directly related to the efficiency of CP predictions. Henceforth, we denote the function $\beta(\delta,n_{\text{tr}})$ as $\beta$ for brevity.
    \begin{theorem}
		\label{th:thm1}
		Under Assumption \ref{ass:nc} and Assumption \ref{ass:gen_bound}, the  expected set size of the probabilistic CP predictor \eqref{eq:cp_predictor} satisfies the inequality
		\begin{align}
			\frac{\Lambda^{\text{CP}}(Q|\mathcal{D}_{\text{tr}})}{|\mathcal{Y}|}\hspace{-0.1em}\leq &\hspace{-0.3em}\int^{R_{\text{max}}}_{R_{\text{min}}}\hspace{-1em}e^{-n_{\text{cal}} \text{d}_{\text{KL}}\left(\hspace{-0.2em}\frac{n_{\alpha}}{n_{\text{cal}}}\big|\big|\hat{F}(r|Q,\mathcal{D}_{\text{tr}})- \frac{\beta}{\sqrt{n_{\text{tr}}}}\hspace{-0.2em}\right)} \gamma(r|Q,\mathcal{D}_{\text{tr}})\mathrm{d}r\nonumber \\
			&+ \gamma(R_{\text{min}}|Q,\mathcal{D}_{\text{tr}})R_{\text{min}}\label{eq:final}
		\end{align}
        with probability $1-\delta$ with respect to the random draw of the training data set $\mathcal{D}_{\text{tr}}$, 
		where $\text{d}_{\text{KL}}\left(a||b\right)=a\log(a/b)+(1-a)\log((1-a)/(1-b))$ is the binary Kullback-Leibler divergence and we have defined
        \begin{align}
		R_{\text{min}}=\inf_{r\in[0,R_{\text{max}}]}\left\{r:\hat{F}(r|Q,\mathcal{D}_{\text{tr}})\geq \frac{n_{\alpha}}{n_{\text{cal}}}+\frac{\beta}{\sqrt{n_{\text{tr}}}} \right\}.
		\label{eq:rmin}
	\end{align}
	\end{theorem}

	\begin{figure}
		\centering
		\includegraphics[width=0.45\textwidth]{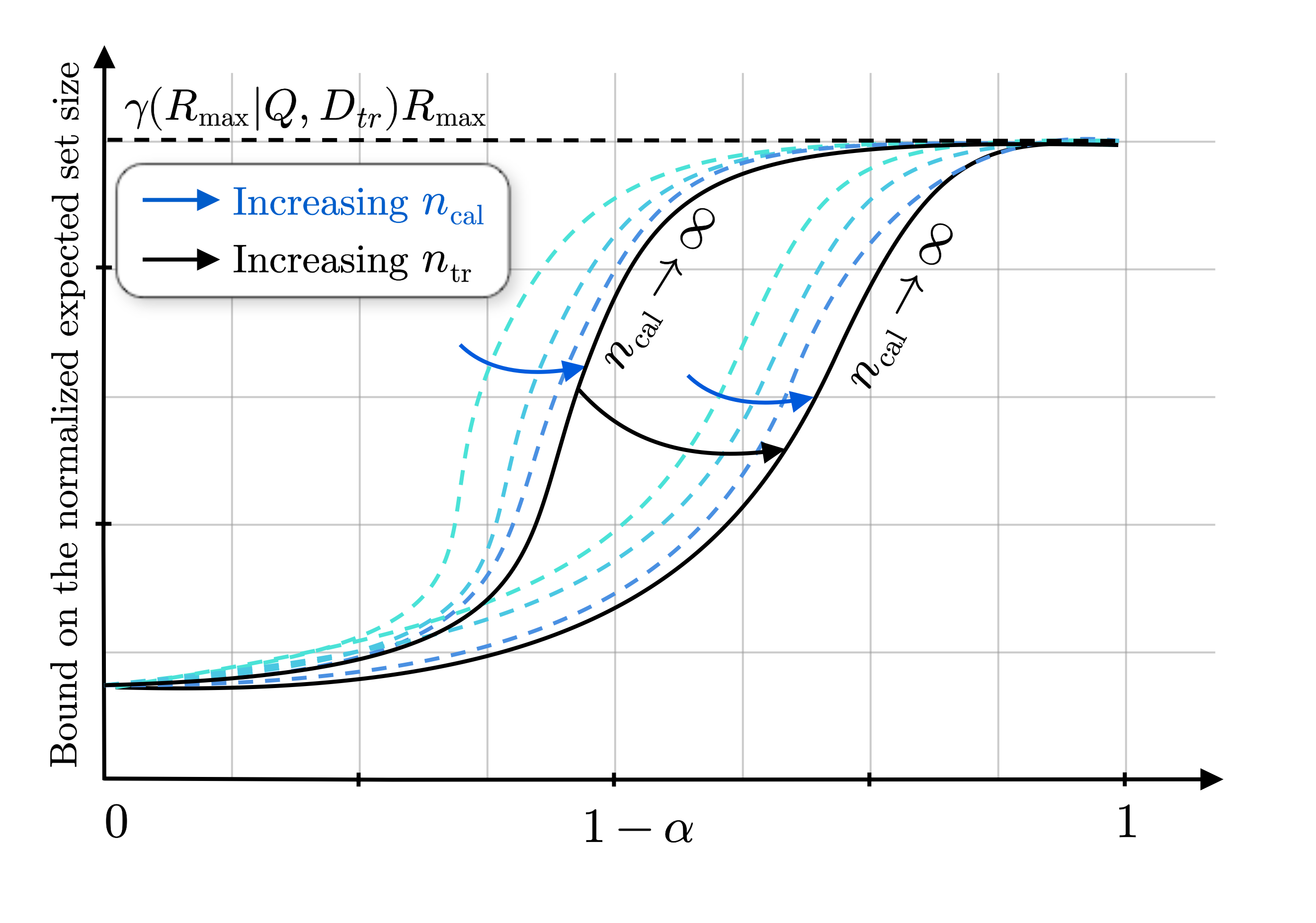}
         \vspace{-1em}
		\caption{Bound on the average set size \eqref{eq:final} for different values of $n_{\text{tr}}$ and $n_{\text{cal}}$ as a function of the target reliability level $1-\alpha$. Increasing  the number $n_{\text{cal}}$ of calibration data points causes the bound to converge exponentially fast to a function (black line) that is  increasing in $1-\alpha$ and decreasing in the amount of training data $n_{\text{tr}}$. }
		\label{fig:evolution}
   \vspace{-1em}
	\end{figure}

 \subsection{Discussion}
	The bound \eqref{eq:final} on the expected size of the CP set predictor presented in Theorem \ref{th:thm1} offers insights into the efficiency of CP as a function of the  generalization performance of the predictor, which is accounted for by the function $\beta>0$ and by the size of the training set $n_{\text{tr}}$ via Assumption \ref{ass:gen_bound}, as well as of the calibration data set size $n_{\text{cal}}$ and the target reliability level $1-\alpha$. 
 
 To start, we observe that the first term in \eqref{eq:final} decreases exponentially fast with an increase in the amount of calibration data $n_{\text{cal}}$. Consequently, for large values of $n_{\text{cal}}$, the bound on the set predictor size can be approximated by the second term, namely $\gamma(R_{\text{min}}|Q,\mathcal{D}_{\text{tr}})R_{\text{min}}$, which is roughly constant with respect to $n_{\text{cal}}$, since we have the approximate equality $n_{\alpha}/n_{\text{cal}}\approx 1-\alpha$ for $n_{\text{cal}}$ sufficiently large. Overall, as illustrated in Figure \ref{fig:evolution}, as the amount of calibration data $n_{\text{cal}}$ grows, the bound \eqref{eq:final} on the efficiency of the CP set size converges exponentially fast to a quantity, $\gamma(R_{\text{min}}|Q,\mathcal{D}_{\text{tr}})R_{\text{min}}$,  drawn as a black line, that is increasing in the target reliability level $1-\alpha$ and decreasing in the training set size $n_{\text{tr}}$.

	 For a finite amount of calibration data $n_{\text{cal}}$, both terms in \eqref{eq:final} are generally non-negligible. In this regard, a key observation is that the exponent  $\text{d}_{\text{KL}}(n_{\alpha}/n_{\text{cal}}\big|\big|\hat{F}(r|Q,\mathcal{D}_{\text{tr}})- \beta/\sqrt{n_{\text{tr}}})$ that dictates the  decrease rate of the first term grows with the discrepancy between the target reliability level $1-\alpha$ -- more precisely, the fraction $ n_{\alpha}/n_{\text{cal}}\approx1-\alpha$ of calibration NC scores included in the CP predicted set -- and the corrected empirical training c.d.f. $\hat{F}(r|Q,\mathcal{D}_{\text{tr}})- \beta/\sqrt{n_{\text{tr}}}$. Thus, the exponent tends to increase as the target reliability level $1-\alpha$ decreases, indicating smaller sets for less demanding reliability targets.

  The exponential decrease rate of the first term in \eqref{eq:final} justifies the common practice of allocating a larger amount of training data $n_{\text{tr}}$ to train the base model compared to the size of the calibration set $n_{\text{cal}}$. However, as sketched in Figure \ref{fig:evolution}, the interplay between these two decreasing terms is not straightforward, as one needs to ensure that enough data points are allocated to calibration to control the average set size.

	\section{Examples}
	In this section, we provide two representative examples, one for classification and one for regression, with the main goal of instantiating the general bound in Theorem \ref{th:thm1} and of comparing the numerical evaluations of the expected set size with the behavior predicted by Theorem \ref{th:thm1}.
	\subsection{Multi-class Classification}
    
	Consider the multi-class classification problem with $|\mathcal{Y}|$ classes, where we choose as NC score the standard 0-1 loss
	\begin{align}
		R(f_\theta(x),y)=\mathds{1}\left\{y\neq f_\theta(x)\right\}.
	\end{align}
	Using \eqref{eq:cp_predictor}, for any reliability level $1-\alpha$, the associated CP predictor is 
	\begin{align}
		\Gamma^{\text{CP}}_{\theta}(X|\mathcal{D}_{\text{cal}})=\begin{cases}
			\{f_\theta(X)\}, &\text{ if }  \mathcal{Q}_{1-\alpha}(\mathcal{R}_{\text{cal}})=0\\
			\mathcal{Y}, &\text{ if } \mathcal{Q}_{1-\alpha}(\mathcal{R}_{\text{cal}})=1,
		\end{cases}
	\end{align}
    that is, the predicted set includes the entire label set $\mathcal{Y}$ if the number of correctly classified calibration points is less than $n_{\alpha}$, and it includes only the point prediction $f_\theta(X)$ otherwise.
	
 \subsubsection{Evaluating the bound}The corresponding size  \eqref{eq:exp_mf} of the NC score $r$  is \eqref{eq:scoreclass}. Accordingly, the empirical training  c.d.f. is given by $\hat{F}(0|Q,\mathcal{D}_{\text{tr}})=0$ and 
	\begin{align}
		\hat{F}(1|Q,\mathcal{D}_{\text{tr}})=\frac{1}{n_{\text{tr}}}\sum^{n_{\text{tr}}}_{i=1} \Pr[f_{\theta}(X_i)=Y_i|\mathcal{D}_{\text{tr}}]=\hat{P}_{\text{tr}},
	\end{align}
    where $\hat{P}_{\text{tr}}$ is the expected average fraction of training data points that are correctly classified.
	Evaluating the bound \eqref{eq:final}, we obtain that, if $\hat{P}_{\text{tr}}\geq n_{\alpha}/n_{\text{cal}}+ \beta/{\sqrt{n_{\text{tr}}}}$, i.e., if the base predictor has a sufficiently high training accuracy, then the normalized expected set size satisfies the inequality
	\begin{align}	\!\!\!\frac{\Lambda^{\text{CP}}(Q|\mathcal{D}_{\text{tr}})}{|\mathcal{Y}|}
 \!\leq\!
		\frac{1}{|\mathcal{Y}|} \!+\! \left(1 \!-\! \frac{1}{{|\mathcal{Y}|}}\right) e^{-n_{\text{cal}} \text{d}_{\text{KL}}\left(\frac{n_{\alpha}}{n_{\text{cal}}}\big|\big|\hat{P}_{\text{tr}}- \frac{\beta}{\sqrt{n_{\text{tr}}}}\right)}.
		\label{01loss}
	\end{align}
	
By the bound (\ref{01loss}), increasing the number $n_{\text{cal}}$ of calibration data points reduces the magnitude of the second term in (\ref{01loss}) at an exponential rate, until the expected set-size reaches the minimum normalized size $1/{|\mathcal{Y}|}$. Furthermore, the rate of decrease of the prediction set size grows as the largest reliability level $1-\alpha$ becomes smaller than the corrected empirical training estimate $\hat{P}_{\text{tr}}- {\beta}/{\sqrt{n_{\text{tr}}}}$. Finally, when the accuracy of the predictor is  insufficient, i.e., when $\hat{P}_{\text{tr}}<n_{\alpha}/n_{\text{cal}}+ {\beta}/{\sqrt{n_{\text{tr}}}}$, the bound (\ref{01loss}) on the expected set size becomes vacuous, yielding $	\Lambda^{\text{CP}}(Q|\mathcal{D}_{\text{tr}})/|\mathcal{Y}|=1$.
	
	\subsubsection{Numerical results}
	\label{sec:bin_sim}
\begin{figure}
		\centering
		\includegraphics[width=0.48\textwidth]{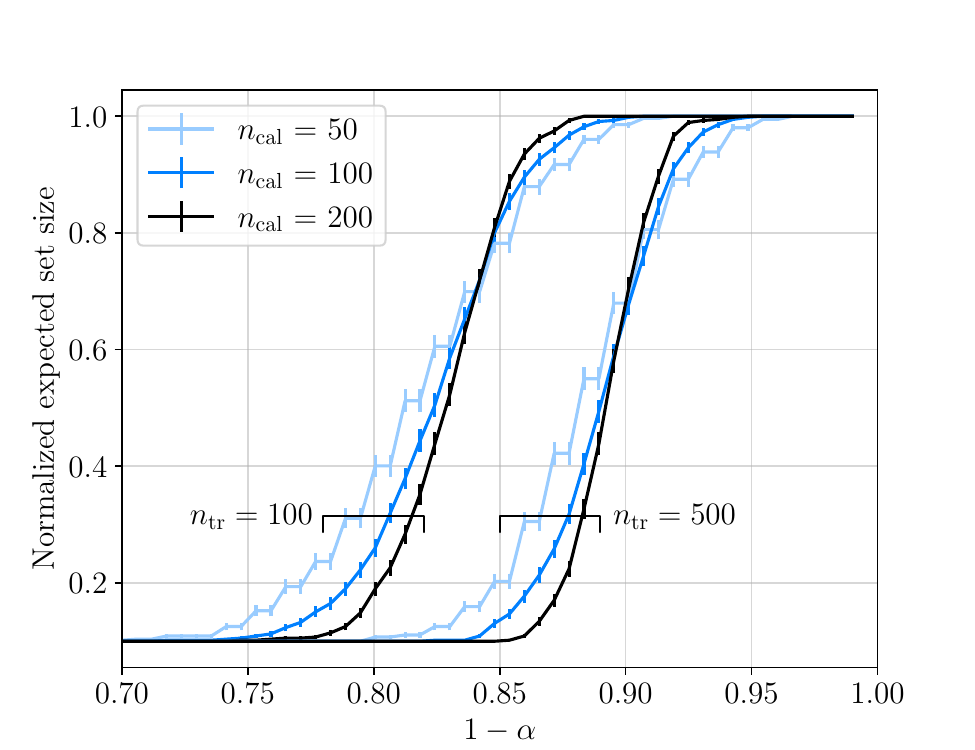}
		\caption{Normalized empirical CP set size for a multi-class classification problem on the MNIST data set as a function of the reliability level $1-\alpha$ and for different sizes of the calibration and training data sets.   }
		\label{fig:classification}
         \vspace{-0.7em}
	\end{figure}

	We empirically validate the behavior of the expected set size by studying a simple classification task. Specifically, we consider the problem of classifying hand-written digits from the MNIST dataset, which has $|\mathcal{Y}|=10$ classes.  We adopt set predictors obtained by calibrating two different logistic regression models, one trained on a data set of size $n_{\text{tr}}=100$ and one trained on a larger data set of size  $n_{\text{tr}}=500$. The resulting classifiers are calibrated using calibration data sets of increasing size $n_{\text{cal}} \in \{50,100,200\}$. 
	In Figure \ref{fig:classification}, we report the normalized CP prediction set size $\Lambda^{\text{CP}}(Q|\mathcal{D}_{\text{tr}})/|\mathcal{Y}|$ for different reliability values $1-\alpha$. 
 
 Comparing the empirical results in Figure \ref{fig:classification} with the theoretical sketch derived from our theory in Figure \ref{fig:evolution} confirms the validity of the general insights presented in the previous section. First, when increasing the size $n_{\text{cal}}$  of the calibration data set, the performance converges to a function of the reliability $1-\alpha$ (black line) that increases with the training set size $n_{\text{tr}}$. In this regard, a minor caveat is that, unlike the prediction of the bound \eqref{eq:final}, there is a small range of reliability levels in which the  expected set size tends to this function from the right rather than from the left. 
 
	\subsection{Example: $\ell_p$ Regression}
     
	We now consider a one-dimensional regression problem with bounded target domain $\mathcal{Y}=[-B_u,B_l]$ and with the NC score being the $\ell_p$-norm
	\begin{align}
		R(x,y)=\lVert x-y\rVert_p.
	\end{align}
	Based on \eqref{eq:cp_predictor}, the corresponding CP set predictor is given  by
	\begin{align}
		\Gamma^{\text{CP}}_{\theta}(X|\mathcal{D}_{\text{cal}})\!=\!\left\{y\in \mathcal{Y}: \lVert f_{\theta}(X)-y\rVert_p \!\leq\! \mathcal{Q}_{1-\alpha}(\mathcal{R}_{\text{cal}})\right\},
	\end{align}
    i.e., it includes all values of $\mathcal{Y}$ with an $\ell_p$ distance from the point prediction $f_{\theta}(X)$ that is no larger than $\mathcal{Q}_{1-\alpha}(\mathcal{R}_{\text{cal}})$.
	\subsubsection{Evaluating the bound} For the $\ell_p$ NC score, the  score size \eqref{eq:exp_mf} can be computed in closed form as in \eqref{eq:scoreregr}. 
    The empirical training NC score c.d.f. is given by 
	\begin{align}
		\hat{F}(r|Q,\mathcal{D}_{\text{tr}})=\frac{1}{n_{\text{tr}}}\sum^{n_{\text{tr}}}_{i=1} \Pr[\lVert f_{\theta}(X_i)-Y_i\rVert_p<r|\mathcal{D}_{\text{tr}}],
	\end{align}
    which corresponds to the expected fraction of training samples with an $\ell_p$ error less than $r$. Instantiating the bound in \eqref{eq:final}, we find that the normalized expected CP satisfies the inequality
	\begin{align}
		\frac{\Lambda^{\text{CP}}(Q|\mathcal{D}_{\text{tr}})}{B_u-B_l}\hspace{-0.2em}\leq &\hspace{-0.3em}\int^{R_{\text{max}}}_{R_{\text{min}}}\hspace{-1em}e^{-n_{\text{cal}} \text{d}_{\text{KL}}\left(\hspace{-0.2em}\frac{n_{\alpha}}{n_{\text{cal}}}\big|\big|\hat{F}(r|Q,\mathcal{D}_{\text{tr}})- \frac{\beta}{\sqrt{n_{\text{tr}}}}\hspace{-0.2em}\right)}\hspace{-0.2em}\frac{2r^{1/p-1}}{p(B_u-B_l)}dr\nonumber \\
		&+\frac{2R_{\text{min}}^{1/p}}{p(B_u-B_l)}.
		\label{eq:lp}
	\end{align}

	\subsubsection{Numerical results}
	To validate the analysis, we study a one-dimensional regression problem based on the California housing data set. To this end, we train a two-hidden-layer perceptron with 50 neurons per layer to predict the value of a property based on an $8$ dimensional feature vector. We consider two neural networks, one trained using $n_{\text{tr}}=100$ samples and one trained using $n_{\text{tr}}=500$ samples. The resulting models are calibrated using the $\ell_p$ NC score with $p=2$ and with different calibration data set sizes $n_{\text{cal}} \in \{50,100,200\}$. 
 
\begin{figure}
		\centering
		\includegraphics[width=0.48\textwidth]{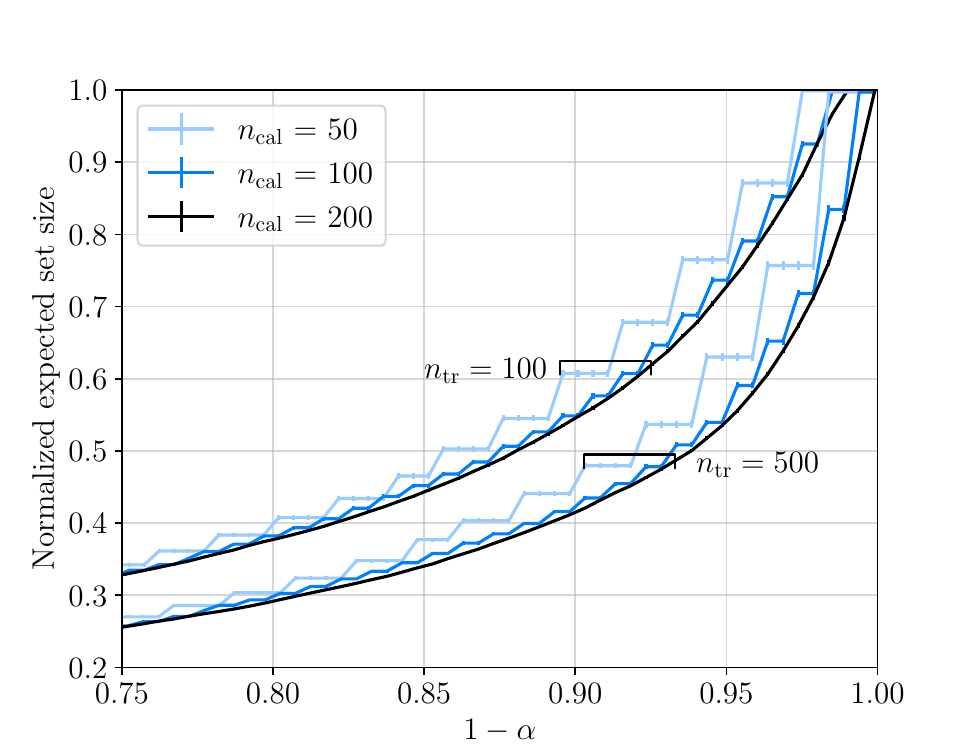}
		\caption{Normalized empirical CP set size for an $\ell_p$ regression task with $p=2$ on the California housing data set as a function of the reliability level $1-\alpha$ and for different sizes of the calibration and training data sets.   }
		\label{fig:regression}
   \vspace{-0.7em}
	\end{figure}

 In Figure \ref{fig:regression}, we again show the normalized empirical CP set size $\Lambda^{\text{CP}}(Q|\mathcal{D}_{\text{tr}})/|\mathcal{Y}|$ as a function of the reliability threshold $1-\alpha$. Like the classification example, the results confirm the general conclusions produced by our theory. In particular, increasing the calibration data set size $n_{\text{cal}}$ yields a prediction set size that decreases with the number of data points in the training set  (black lines). 

	\section{Conclusion}
    In this work, we have investigated the informativeness of set predictors produced by CP by providing an upper bound on the expected predicted set size. The bound illustrates the interplay between informativeness of the prediction and the generalization properties of the base predictor. An interesting research direction may attempt to translate the derived   bound into a practical training algorithm to produce efficient set predictors with coverage guarantees \cite{park2023few,stutz2021learning}.

     \section*{Acknowledgments}
This work was partially supported by the European Union’s Horizon Europe project CENTRIC (101096379), by the Open Fellowships of the EPSRC (EP/W024101/1), by the EPSRC project (EP/X011852/1), by the UK Government under Project REASON, and by the Wallenberg AI, Autonomous Systems and Software Program (WASP) funded by the Knut and Alice Wallenberg Foundation.
	\newpage
	\bibliography{biblio}
	\bibliographystyle{ieeetr}

	\newpage
        \newpage

        \twocolumn
 
	\appendix

\subsection{On Assumption \ref{ass:gen_bound} }
 \label{app:lemma}

In this section, we establish that Assumption \ref{ass:gen_bound} holds for several common learning algorithms.
To this end, we let $P$ denote the marginal distribution on the model parameter space $\Theta$ induced by the product distribution $Q\times P_Z^n$ on  $\Theta\times (\mathcal X \times \mathcal Y)^n$.
With this definition, the mutual information between the model parameter and the training data set can be expressed as the average  $I(\theta;\mathcal D_{ \text{tr} })=\mathbb{E}_{\mathcal D_{ \text{tr} }} [\text{KL}(Q||P)]$. The key observation is that for Gibbs posteriors, $\epsilon$-differentially private algorithms with $\epsilon=O(\log(n_{\text{tr}})/n_{\text{tr}})$, and stochastic gradient Langevin dynamics (SGLD), the mutual information scales as \cite{pensia-18a,feldman-18a,raginsky-21a} \begin{equation}\label{eq:mutinfo}I(\theta;\mathcal D_{ \text{tr} })=O(\log \ntr). \end{equation} 

In the following lemma, which is a direct consequence of Markov's inequality, we show that, if  the mutual information is upper bounded by a logarithmic function of  the number of training points, $\ntr$, as per (\ref{eq:mutinfo}), this implies that the Kullback-Liebler (KL) divergence $\text{KL}(Q||P)$ is also logarithmic in  $\ntr$ with high probability.

\begin{lemma}\label{lem:kl-bound-high-prob}
    Assume that $Q$ is absolutely continuous with respect to $P$ and that the inequality $I(\theta;\mathcal D_{ \text{tr} })\leq c\log (\ntr)$ holds for some constant $c>0$.
    Then, for any $\delta\in(0,1)$,  with probability at least $1-\delta$, we have the following inequality
    \begin{align}\label{eq:bound-app}
       \text{KL}(Q||P) \leq \frac{I(\theta;\mathcal D_{ \text{tr} })}{\delta} \leq \frac{ c\log(\ntr) }{\delta} .
    \end{align}
\end{lemma}

In the following proposition, we show that Lemma~\ref{lem:kl-bound-high-prob} suffices to motivate Assumption~\ref{ass:gen_bound} for the listed learning algorithms.

\begin{proposition}
If the condition in (\ref{eq:bound-app}) is satisfied, Assumption~\ref{ass:gen_bound} holds.\end{proposition}

\begin{proof}
	As in (\ref{eq:cal_cdf}), we define the empirical training c.d.f. at the NC score level $r$ for a fixed model $\theta$ as
	\begin{align}
		\hat{F}(r|\theta,\mathcal{D}_{\text{tr}})=\frac{1}{n_{\text{tr}}}\sum_{i=1}^{n_{\text{tr}}}\mathds{1}\left\{R(f_{\theta}(x_i),y_i)< r \right\},
	\end{align} 
	as well as the corresponding true population c.d.f. as in (\ref{eq:true_pop_cdf})
	\begin{align}
		F(r|\theta)=	\mathbb{E}_{X,Y}\left[\mathds{1}\left\{R(f_{\theta}(X),Y)< r \right\}\right].
	\end{align} 
	For a given model $\theta$, we denote the maximum absolute discrepancy between the population c.d.f. and the empirical training c.d.f. as in (\ref{eq:cdf_errror}), i.e.,
	\begin{align}
		\Delta(\theta|\mathcal{D}_{\text{tr}})=\sup_{r\in [0,R_{\text{max}}]}|F(r|\theta)-\hat{F}(r|\theta,\mathcal{D}_{\text{tr}})|.
	\end{align}
	Following a symmetrization argument \cite[Lemma~1.1]{case1984bounded}, we have the inequality  
	\begin{align}
		\mathbb{E}_{\mathcal{D}_{\text{tr}}}[e^{\lambda{	\Delta(\theta|\mathcal{D}_{\text{tr}})}}]\leq e^{\frac{2\lambda^2}{n_{\text{tr}}}\log(4)}, \quad \forall \lambda\geq 0. \label{eq:olivera_imbuzeiro}
	\end{align}
 Thus, the random variable $\Delta(\theta|\mathcal{D}_{\text{tr}})$ is $2\sqrt{\log(4)/n_{\text{tr}}}$-sub-Gaussian~\cite[Thm.~2.6]{wainwright-19a}.
	
 Next, the Donsker and Varadhan identity \cite{alquier2021user} states that for a measurable and bounded function $h(\theta)$, we have 
	\begin{align}
		\log \left[\mathbb{E}_{\theta\sim P}\left[ e^{h(\theta)}\right]\right]\!=\!\!\!\sup_{Q\in\mathcal{P}(\Theta)}\left[\mathbb{E}_{\theta\sim Q}[h(\theta)]-\text{KL}(Q||P)\right],
		\label{eq:dv_ineq}
	\end{align}
	where $\mathcal{P}(\Theta)$ is the set of all probability distributions on $\Theta$.
 Applying (\ref{eq:dv_ineq}) to $(n_{\text{tr}}-1)\Delta(\theta|\mathcal{D}_{\text{tr}})^2/(16\log(2))$, we obtain
	\begin{align}
		&\mathbb{E}_{\theta\sim P}\mathbb{E}_{\mathcal{D}_{\text{tr}}}[e^{\frac{(n_{\text{tr}}-1)\Delta(\theta|\mathcal{D}_{\text{tr}})^2}{16\log(2)}}]\\
  &=\mathbb{E}_{\mathcal{D}_{\text{tr}}}\mathbb{E}_{\theta\sim P}[e^{\frac{(n_{\text{tr}}-1)\Delta(\theta|\mathcal{D}_{\text{tr}})^2}{16\log(2)}  }]\\&=\mathbb{E}_{\mathcal{D}_{\text{tr}}}[e^{\sup_{Q}\frac{n_{\text{tr}}-1}{16\log(2)}\mathbb{E}_{\theta\sim Q}[ \Delta(\theta|\mathcal{D}_{\text{tr}})^2	]-\text{KL}(Q||P)}]\\
  &\leq \sqrt{n_{\text{tr}}},
		\label{dvineq}
		\end{align}
		where the first equality follows from Fubini's Theorem and the inequality follows by \cite[Thm.~2.6.(IV)]{wainwright-19a} with $\lambda=1-1/n_{\text{tr}}$.
 From the Chernoff bound, we deduce the inequality
	\begin{align}
		\label{eq:almost_there_start}&\Pr_{\mathcal{D}_{\text{tr}}}\bigg[
 \sup_{Q\in\mathcal{P}(\Theta)}
 \frac{n_{\text{tr}}-1}{16\log(2)}\mathbb{E}_{\theta\sim Q}[ \Delta(\theta|\mathcal{D}_{\text{tr}})^2]-\hspace{-0.2em}\text{KL}(Q||P)\hspace{-0.2em}
 >s\bigg] \nonumber \\
 &\leq\mathbb{E}_{\mathcal{D}_{\text{tr}}}[e^{
 \sup_{Q\in\mathcal{P}(\Theta)}
 \frac{n_{\text{tr}}-1}{16\log(2)}\mathbb{E}_{\theta\sim Q}[ \Delta(\theta|\mathcal{D}_{\text{tr}})^2]-\text{KL}(Q||P)
 }]e^{-s}\\
 &\leq e^{\log(\sqrt{\ntr})-s},
		\label{eq:almost_there}
	\end{align} 
	where the last inequality follows from (\ref{dvineq}).
	
 Rearranging terms in (\ref{eq:almost_there_start})--(\ref{eq:almost_there}), we get
	\begin{align}&\Pr_{\mathcal{D}_{\text{tr}}}\bigg[\sup_{Q\in\mathcal{P}(\Theta)}\mathbb{E}_{\theta\sim Q}[{	\Delta(\theta|\mathcal{D}_{\text{tr}})}^2]\nonumber\\
 &\qquad>\frac{16\log(2)(\text{KL}(Q||P)+s+\log\sqrt{n_{\text{tr}}})}{n_{\text{tr}}-1}\bigg]\leq e^{-s}.
	\end{align} 
	By Jensen's inequality, it follows that $\mathbb{E}_{\theta\sim Q}[ \Delta(\theta|\mathcal{D}_{\text{tr}})^2]\geq (\mathbb{E}_{\theta\sim Q}[ \Delta(\theta|\mathcal{D}_{\text{tr}})])^2$. Furthermore, we have that
	\begin{align}
		\mathbb{E}_{\theta\sim Q}	\left[\Delta(\theta|\mathcal{D}_{\text{tr}})\right]&=\mathbb{E}_{\theta\sim Q}\left[\sup_{r\in\mathbb{R}}|F(r|\theta)-\hat{F}(r|\theta,\mathcal{D}_{\text{tr}})|\right]\\
  &\geq\sup_{r}\mathbb{E}_{\theta\sim Q}\left[|F(r|\theta)-\hat{F}(r|\theta,\mathcal{D}_{\text{tr}})|\right]\\
  &\geq \Delta(Q|\mathcal{D}_{\text{tr}}).
	\end{align}
	The final result is obtained by choosing $s=-\log(\delta)$, which leads to 
   \begin{align}
\Pr_{\mathcal{D}_{\text{tr}}}&\left[\sup_{Q\in\mathcal{P}(\Theta)}\Delta(Q|\mathcal{D}_{\text{tr}})\leq 4\sqrt{\log 2}\sqrt{\frac{\text{KL}(Q||P)+\log\frac{\sqrt{n_{\text{tr}}}}{\delta}}{{n_{tr}}-1}}\right]\nonumber\\
  &\geq 1-\delta.
		\label{eq:final_app}
	\end{align} 
 Using the union bound, we combine this with the assumed bound (\ref{eq:bound-app}), with the inequality $n_{\text{tr}}-1\leq 2n_{\text{tr}}$, and we let $\delta \rightarrow \delta/2$.
 Thus, Assumption \ref{ass:gen_bound} holds with the function
 \begin{equation}
     \beta(\delta,n_{\mathrm{tr}}) = \sqrt{32\log 2\left(\frac{2c\log(n_{\text{tr}})}{\delta} + \log \frac{2\sqrt{ n_{ \text{tr} } }}{\delta} \right) }.
 \end{equation}
\end{proof}


    \subsection{Proof of Theorem \ref{th:thm1}}
	The expected set size is 
	\begin{align}
		\Lambda^{\text{CP}}(Q|\mathcal{D}_{\text{tr}})=\mathbb{E}\left[|\Gamma^{\text{CP}}_\theta(X|\mathcal{D}_{\text{cal}})|\right],
	\end{align}
	where the expectation is taken over the calibration data $\mathcal{D}_{\text{cal}}\sim P_{Z}^{\otimes n_{\text{cal}}}$, the input $X\sim P_X$, and the model $\theta\sim Q(\theta|\mathcal{D}_{\text{tr}})$.
	For a reliability level $\alpha$, the expected set size is \cite{dhillon2023expected}
	\begin{align}
		\Lambda^{\text{CP}}&(Q|\mathcal{D}_{\text{tr}})\hspace{-0.25em}\\
  =&\mathbb{E}\left[\int_y\mathds{1}\left\{y\in \mathcal{Y}\hspace{-0.25em}:\hspace{-0.25em}R(f_\theta(X),y)\hspace{-0.2em}\leq\hspace{-0.25em}\mathcal{Q}_{1-\alpha}(\mathcal{R}_{\text{cal}}) \right\}dy\right]\\
	=&\mathbb{E}\left[\int_y\Pr\left[R(f_\theta(X),y)\leq\mathcal{Q}_{1-\alpha}(\mathcal{R}_{\text{cal}})|D_{\text{tr}}\right]dy\right]\\
		=&\mathbb{E}\left[\int_y\Pr\left[\mathcal{Q}_{1-\alpha}(\mathcal{R}_{\text{cal}}) \!\geq\! r \big| R(f_\theta(X),y)\!=\!r,D_{\text{tr}}\right]dy\right]\\
  &+\mathbb{E}\left[\int_y\Pr\left[R(f_\theta(X),y)= r \big|D_{\text{tr}} \right]dy\right] \nonumber\\
		=&\int_{\mathcal{R}}\Pr\left[\mathcal{Q}_{1-\alpha}(\mathcal{R}_{\text{cal}})\geq r\big|D_{\text{tr}}\right]\gamma(r|Q,\mathcal{D}_{\text{tr}})|\mathcal{Y}| dr,
		\label{eq:expected}
	\end{align}
	where the last equality follows from the independence of $\mathcal{R}_{\text{cal}}$ and $R(f_\theta(X),y)$, and the definition of the weighting factor $\gamma(r|Q,\mathcal{D}_{\text{tr}})$ in \eqref{eq:exp_mf}. 
 
 By the definition of $\mathcal{Q}_{1-\alpha}(\mathcal{R}_{\text{cal}})$, we have the equivalence
	\begin{align}
		\!\!\!\mathcal{Q}_{1-\alpha}(\mathcal{R}_{\text{cal}})\geq r  
  \!\iff\! \sum^{n_{\text{cal}}}_{i=1} \mathds{1}\left\{ R(f_{\theta_i}(X_i),Y_i) \!<\! r  \right\} \leq n_\alpha,
	\end{align}
	with $ n_\alpha=\lceil(n_{\text{cal}}+1)(1-\alpha)\rceil-1$. It  then follows that
	\begin{align}
 \!\!\! \Pr\left[\mathcal{Q}_{1\!-\!\alpha}(\mathcal{R}_{\text{cal}}) \!\geq\! r\big|D_{\text{tr}}\right] \!=\! \Pr\left[\sum^{n_{\text{cal}}}_{i=1}  \text{Bern}(p_i(r))\!\leq\!  n_\alpha\bigg|D_{\text{tr}}\!\right] ,\!\!
	\end{align}
	where $\text{Bern}(p_i(r))$ denotes the Bernoulli random variable of parameter $p_i(r)=\Pr[R(f_{\theta_i}(X_i),Y_i)< r]$.
	By the definition of calibration NC score \eqref{eq:cal_NC_score}, the random variables $\{\text{Bern}(p_i(r))\}^{n_{\text{cal}}}_{i=1}$ are independent and identically distributed, and thus we have the equality $p_i(r)=F(r|Q,\mathcal{D}_{\text{tr}})$ for all $i\in {1,\dots,n_{\text{cal}}}$ with $F(r|Q,\mathcal{D}_{\text{tr}})$ defined in (\ref{eq:true_pop_cdf}). Therefore, we have the equality 
	\begin{multline}
		\Pr\left[\mathcal{Q}_{1-\alpha}(\mathcal{R}_{\text{cal}})\hspace{-0.2em}\geq \hspace{-0.2em}r\big|\hspace{-0.1em} D_{\text{tr}}\right]\hspace{-0.2em}\\
  =\hspace{-0.2em}\Pr\left[\text{Bin}(n_{\text{cal}},F(r|Q,\mathcal{D}_{\text{tr}}))\hspace{-0.2em}\leq \hspace{-0.2em}n_\alpha\big|\hspace{-0.1em}D_{\text{tr}}\right],
	\end{multline}
	with $\text{Bin}(n_{\text{cal}},F(r|Q,\mathcal{D}_{\text{tr}}))$ being the binomial random variable with parameters $n_{\text{cal}}$ and $F(r|Q,\mathcal{D}_{\text{tr}})$.  
	
	From Assumption \ref{ass:gen_bound}, with probability $1-\delta$ with respect to the realization of the training data set $\mathcal{D}_{\text{tr}}$, the probability $F(r|Q,\mathcal{D}_{\text{tr}})$ can be lower bounded as 
	\begin{multline}
		F(r|Q,\mathcal{D}_{\text{tr}})\hspace{-0.15em}\\
  \geq\hspace{-0.1em} \underbrace{\frac{1}{n_{\text{tr}}}\hspace{-0.1em}\sum^{n_{\text{tr}}}_{i=1}\mathbb{E}_{Q(\theta|\mathcal{D}_{\text{tr}})}\left[\mathds{1}\left\{R(f_{\theta}(x_i),y_i)< r\right\}\right]}_{=\hat{F}(r|Q,\mathcal{D}_{\text{tr}})}-\frac{\beta}{\sqrt{n_{\text{tr}}}} \hspace{-0.1em}.
        \label{eq:difficult}
	\end{multline}

	Since the binomial c.d.f. $\text{Bin}(n_{\text{cal}},p(r))$ is decreasing in the probability $p(r)$, we have
	\begin{align}
		\Pr&\left[\mathcal{Q}_{1-\alpha}(\mathcal{R}_{\text{cal}})\geq r\big|D_{\text{tr}}\right]&\hspace{-0.2em}\\
&\leq\hspace{-0.2em}\Pr\left[\text{Bin}\left(\hspace{-0.2em}n_{\text{cal}},\hat{F}(r|Q,\mathcal{D}_{\text{tr}})\hspace{-0.2em}-\hspace{-0.2em}\frac{\beta}{\sqrt{n_{\text{tr}}}}\hspace{-0.2em}\right)\hspace{-0.2em}\leq \hspace{-0.2em}n_\alpha\bigg|D_{\text{tr}}\right].
	\end{align}
	Having defined $R_{\text{min}}$ as in \eqref{eq:rmin}, for values of $r\geq R_{\text{min}}$ we have the inequality $\hat{F}(r|Q,\mathcal{D}_{\text{tr}})-\frac{\beta}{\sqrt{n_{\text{tr}}}}\geq \frac{n_{\alpha}}{n_{\text{cal}}}$. From the Chernoff bound, we then have the inequality
	\begin{align}
		\Pr\left[\mathcal{Q}_{1-\alpha}(\mathcal{R}_{\text{cal}})\geq r\big|D_{\text{tr}}\right]\leq e^{-n_{\text{cal}} \text{d}_{\text{KL}}\left(\frac{n_{\alpha}}{n_{\text{cal}}}\Big|\Big|\hat{F}^{\text{tr}}(r)- \frac{\beta+\mu}{\sqrt{n_{\text{tr}}}}\right)}
		\label{eq:chernoff}
	\end{align}
    for $r\geq R_{\text{min}}$, whereas for $r<R_{\text{min}}$ we use the trivial bound
	\begin{align}
		\Pr\left[\mathcal{Q}_{1-\alpha}(\mathcal{R}_{\text{cal}})\geq r\big|D_{\text{tr}}\right]\leq 1.
		\label{eq:vacous}
	\end{align}
	Combining \eqref{eq:vacous} and \eqref{eq:chernoff} into \eqref{eq:expected}, and using the monotonicity of the multiplicative factor $\gamma(r|Q,\mathcal{D}_{\text{tr}})$ (Assumption \ref{ass:nc}), we obtain the desired bound. 
    \subsection{Estimating the Bound}

	The evaluation of the expected set size bound \eqref{eq:final} provided in Theorem \ref{th:thm1} requires calculating the  empirical training c.d.f. $\hat{F}(r|Q,\mathcal{D}_{\text{tr}})$, which includes an expectation with respect to the model $\theta\sim Q(\theta|\mathcal{D}_{\text{tr}})$. To sidestep this practical difficulty, we consider estimating the empirical training c.d.f. using a finite number of draws from the model distribution $Q(\theta|\mathcal{D}_{\text{tr}})$, obtaining an expected set size bound that is easier to evaluate. 
 
    Denoting as $\theta_i\sim Q(\theta|\mathcal{D}_{\text{tr}})$ for $i=1,\dots,n_{\text{tr}}$, the models drawn independently for each data point $i$, we define the \emph{doubly} empirical training NC score c.d.f. as 
	\begin{align}
		\hat{F}_{\theta}(r|Q,\mathcal{D}_{\text{tr}})=\frac{1}{n_{\text{tr}}}\sum_{i=1}^{n_{\text{tr}}}\mathds{1}\left\{R(f_{\theta_i}(x_i),y_i)< r\right\}.
		\label{eq:train_NCscores_emp}
	\end{align}
    We have the following result.
    \begin{corollary}
    \label{th:cor1}
    Given a learning algorithm $Q(\theta|\mathcal{D}_{\text{tr}})$ satisfying Assumption \ref{ass:nc} and Assumption 2, the expected set size of the probabilistic CP predictor \eqref{eq:cp_predictor} satisfies the following inequality
    \begin{align}
        \frac{\Lambda^{\text{CP}}(Q|\mathcal{D}_{\text{tr}})}{|\mathcal{Y}|}\hspace{-0.2em}\leq& \hspace{-0.35em}\int^{R_{\text{max}}}_{R_{\text{min},\theta}}\hspace{-1em}e^{-n_{\text{cal}} \text{d}_{\text{KL}}\left(\hspace{-0.2em}\frac{n_{\alpha}}{n_{\text{cal}}}\big|\big|	\hat{F}_{\theta}(r|Q,\mathcal{D}_{\text{tr}})- \frac{\beta+\mu}{\sqrt{n_{\text{tr}}}}\hspace{-0.2em}\right)} \hspace{-0.1em}\gamma(r|Q,\hspace{-0.05em}\mathcal{D}_{\text{tr}})\hspace{-0.05em}dr \nonumber\\
        &+\gamma(R_{\text{min},\theta}|Q,\mathcal{D}_{\text{tr}})R_{\text{min},\theta}\label{eq:final_emp}
    \end{align}
    with probability $1-2\delta$ with respect to the random draw of the training data set $\mathcal{D}_{\text{tr}}$ and of the models $\{\theta_i\}^{n_{\text{tr}}}_{i=1}$, where we have defined  $\mu=\sqrt{\log(2/\delta)/2}+\sqrt{4\log(n_{\text{tr}}e/2)}$ and
    \begin{align}
		R_{\text{min},\theta}=\inf_{r\in[0,R_{\text{max}}]}\left\{r:\hat{F}_{\theta}(r|Q,\mathcal{D}_{\text{tr}})\geq \frac{n_{\alpha}}{n_{\text{cal}}}+\frac{\beta+\mu}{\sqrt{n_{\text{tr}}}} \right\}.
		\label{eq:rmin_emp}
	\end{align}
	\end{corollary}
	
    \begin{proof}
    		Using McDiarmid's inequality and the fact that decision stumps have VC dimension $d=2$, the value of the function $\hat{F}(r|Q,\mathcal{D}_{\text{tr}})$ can be  {uniformly} bounded based on the estimate \eqref{eq:train_NCscores_emp}. In particular, with probability $1-\delta$, we have that for all $r\in[0,R_{\text{max}}]$ the inequality 
    		\begin{align}
    			\hat{F}(r|Q,\mathcal{D}_{\text{tr}})\geq \hat{F}_{\theta}(r|Q,\mathcal{D}_{\text{tr}})-\frac{\mu}{\sqrt{n_{\text{tr}}}}
    		\end{align}
    		holds, where $\mu=\sqrt{\log(2/\delta)/2}+\sqrt{4\log(n_{\text{tr}}e/2)}$ \cite[Corollary 3.19]{mohri2018foundations}. Defining $R_{\text{min},\theta}$  as in \eqref{eq:rmin_emp}, the final expression is obtained following the same steps as the proof of Theorem~\ref{th:thm1}.
    \end{proof}
\end{document}